\documentclass[runningheads]{llncs}
\usepackage{cite}
\usepackage{amsmath,amssymb,amsfonts}
\usepackage{tabularx}
\usepackage{algorithm}
\usepackage{algpseudocode}
\usepackage{graphicx}
\usepackage{textcomp}
\usepackage{xcolor}
\usepackage[numbers]{natbib}
\usepackage{subfigure}
\usepackage{parskip}
\def\BibTeX{{\rm B\kern-.05em{\sc i\kern-.025em b}\kern-.08em
    T\kern-.1667em\lower.7ex\hbox{E}\kern-.125emX}}
\begin{document}
%
\title{Real-time Motion Planning for Autonomous Vehicles in Dynamic Environments\\}
%



\author{Mohammad Dehghani Tezerjani\inst{1}\orcidID{0009-0003-0790-9831} 
\and
Deyuan Qu\inst{1}\orcidID{0009-0002-5498-3172} 
\and
Sudip Dhakal\inst{1}\orcidID{0009-0003-7685-6753}
\and
Dominic Carrillo\inst{1}\orcidID{0000-0003-2613-5328}
\and
Amir Mirzaeinia\inst{1}\orcidID{0000-0002-7980-6263}
\and
Qing Yang\inst{1}\orcidID{0000-0003-3495-370X}
}

\authorrunning{M. Dehghani T et al.}
\titlerunning{Real-time Motion Planning for AVs in Dynamic Environments}

\institute{University of North Texas, Denton TX 76201, USA 
\email{\{mike.degany,deyuan.qu,sudip.dhakal,dominic.carrillo,amir.mirzaeinia,qing.yang\}@unt.edu}}
\maketitle              

\begin{abstract}
Recent advancements in self-driving car technologies have enabled them to navigate autonomously through various environments. However, one of the critical challenges in autonomous vehicle operation is trajectory planning, especially in dynamic environments with moving obstacles. This research aims to tackle this challenge by proposing a robust algorithm tailored for autonomous cars operating in dynamic environments with moving obstacles. The algorithm introduces two main innovations. Firstly, it defines path density by adjusting the number of waypoints along the trajectory, optimizing their distribution for accuracy in curved areas and reducing computational complexity in straight sections. Secondly, it integrates hierarchical motion planning algorithms, combining global planning with an enhanced $A^*$ graph-based method and local planning using the time elastic band algorithm with moving obstacle detection considering different motion models. The proposed algorithm is adaptable for different vehicle types and mobile robots, making it versatile for real-world applications. Simulation results demonstrate its effectiveness across various conditions, promising safer and more efficient navigation for autonomous vehicles in dynamic environments. These modifications significantly improve trajectory planning capabilities, addressing a crucial aspect of autonomous vehicle technology.


\end{abstract}

\section{Introduction}
Today, the advancement of autonomous vehicle technology holds immense promise, offering
significant benefits such as heightened safety, doubled efficiency, and increased accessibility,
poised to revolutionize transportation systems. A critical aspect of autonomous vehicle development lies in creating a trajectory from the origin to the destination in the streets and
places full of cars and pedestrians, which underscores the importance of high safety to
prevent fatal accidents and damages\cite{PASSMORE2019e272}.
According to the report of NHTSA, in 2021, 42,939 people died due to road accidents. This
number was 39007 in 2020 and 36355 in 2019, which shows the increase in the number of
deaths in recent years\cite{NHTSA}. Most of the fatal cra were caused by drivers' carelessness and
mistakes, and 31\% of these accidents are drunk drivers and 10\% are distracted drivers\cite{singh2015critical, Singh_2018}.
Autonomous vehicles present a groundbreaking opportunity to reduce errors and save lives
significantly. They also serve as essential tools for individuals facing physical or visual
challenges that hinder private car use. Concurrently, statistical data indicates that 86\% of the
American workforce relies on private car transportation, spending an average of 25 minutes
driving daily\cite{bankrateCommutingWork}. By adopting autonomous vehicles, people can make more efficient use of their
time and reduce the likelihood of experiencing neurological disorders linked to driving.
In autonomous driving, three primary components are pivotal: perception, decision-making,
and control\cite{gonzalez2015review}. The decision-making phase, especially critical in car navigation, is dedicated
to generating paths for autonomous vehicles. This task can be segmented into two facets: path
planning without temporal constraints and trajectory planning considering time factors.
Crafting a path with temporal data entails determining factors like the time needed to reach
specific points and the consequent vehicle speed or acceleration. Such planning also
necessitates considering vehicle dynamics, dynamic obstacles, and environmental alterations
not initially factored into the original mapping process\cite{kunchev2006path}.
Achieving autonomous navigation in vehicles relies on accurately describing the environment
through mapping, identifying obstacles within it, generating obstacle-free routes from the
starting point to the destination, and subsequently adhering to these routes. However, a
significant challenge lies in enabling vehicles to navigate through dynamic environments.
While current obstacles in trajectory generation primarily revolve around intricate real-time
calculations in dynamic settings, this study specifically focuses on real-time motion planning
in environments where dynamic obstacles are present.Motion planning strategies can be categorized into four main groups: graph search-based
algorithms, sampling-based algorithms, interpolation curve algorithms, and numerical
optimization methods\cite{gonzalez2015review}.
Graph-based algorithms and sampling-based algorithms are commonly employed for global
planning purposes\cite{sanchez2021path}. While sampling-based algorithms offer faster performance, they are often
probabilistic in nature and may not yield consistent results across iterations. Consequently,
this study adopts an enhanced A* graph-based algorithm for global planning. Moreover,
recent advancements in the field have introduced several algorithms tailored for local
planning to mitigate collisions with moving obstacles. These include the artificial potential
field algorithm \cite{khatib1986real}, the dynamic window approach \cite{fox1997dynamic}, the elastic band algorithm \cite{quinlan1995real}, and the timed elastic band algorithm \cite{rosmann2012trajectory}.
The well-established elastic band method \cite{quinlan1995real} is a dynamic trajectory planning approach that
adapts the path's shape in real-time. Internal forces, predefined within the method, ensure
path continuity, while external forces help navigate around obstacles. However, the
traditional elastic band method lacks consideration for time-related data and dynamic
constraints. Building upon this, a method introduced in \cite{rosmann2012trajectory} extends the classic elastic band
approach to incorporate temporal information in two stages. Initially, discrete intermediate
waypoints are strategically positioned away from obstacles, followed by employing a
dynamic movement model to ensure path continuity.
In \cite{kurniawati2007path}, these two stages are integrated to streamline the process. In contrast to the previously mentioned method, alternative approaches such as those utilized in Autoware's waypoint generation process extract waypoints from the 3D map created during mapping and store them in a file \cite{sudipdhakal}. While this file acts as a guide for path tracking, it operates independently of real-time constraints, thus introducing its own drawbacks. Trajectory optimization
tasks often entail extensive computational efforts, posing challenges in integrating planning
components with control units. Consequently, researchers seek efficient solutions for
trajectory optimization problems. The dynamic window algorithm, introduced in \cite{fox1997dynamic}, samples
circular trajectories within a search space constrained by permissible linear and rotational
speed orders to generate motion. In \cite{lau2009kinodynamic, sprunk2011online}, spline paths are continually optimized
considering dynamic constraints. Additionally, \cite{rosmann2012trajectory} proposes the time elastic band method, an
enhancement of the elastic band method that incorporates time data to optimize trajectories
while avoiding obstacles.
This study further improves the time elastic band method, enabling dynamic adjustments to
trajectory accuracy based on current conditions. Furthermore, it estimates the trajectories of
moving obstacles in the environment using the Kalman filter \cite{welch1995introduction}. A combination of local planning algorithms and a predefined global planner facilitates real-time planning for
autonomous vehicles in dynamic environments. Initially, motion planning in dynamic
environments is conducted without temporal information. As the vehicle progresses along
this path, it identifies new obstacles and adjusts the path accordingly.
Through a hierarchical approach that integrates planning algorithms, this research aims to
design optimal trajectories that minimize travel time while avoiding both stationary and
moving obstacles, while adhering to the vehicle's non-holonomic constraints\cite{rosmann2012trajectory}.




\section{Methodology}
In this study, the global planning method employed is the $A^*$ algorithm, enhanced with a gradient descent optimizer. Initially, the $A^*$ algorithm is applied to assign values to all map grids using a function, effectively creating an analogous discrete potential field within the environment. Subsequently, the gradient descent method is utilized to extract a favourable path from the origin to the destination based on these values. For local planning tasks, the timed elastic band method is preferred due to its numerous advantages over alternative motion planning approaches.

\subsection{Proposed algorithm for global planning}
The proposed approach for global planning in this research involves utilizing the $A^*$
algorithm to assess a portion of the map and employing a decreasing gradient optimizer to
determine the path. The $A^*$ algorithm is adept at finding the shortest path based on the
distance criterion from the goal, making it suitable for global planning in both structured and
unstructured environments. Incorporating heuristic criteria in this method significantly
reduces computational complexity. By employing this approach, higher-probability path
segments within the map are prioritized, effectively creating a representation akin to a
discrete potential field in the environment. Subsequently, the gradient descent method is
employed to derive the shortest path from the starting point to the destination. The $A^*$
algorithm leverages heuristic criteria to streamline calculations, while the use of gradients
facilitates the discovery of smoother paths, enhancing its comparative value over other
planning methods.The implementation of the $A^*$ algorithm in this research is outlined in Algorithm \ref{Alg1}.
\subsection{$A^*$ planner}
In the global planning method applied to the generated map, as integrated within the
$A^*$ algorithm, the procedure entails receiving the cost map coordinates
corresponding to the starting and ending points, and subsequently deriving the path
from the end point. Initially, the path finder initializes the path array by placing the
destination point. It then examines the eight neighbouring cell, selecting the one with
the lowest value as the subsequent waypoint along the path. This iterative process
persists until the starting point is reached.

\subsection{Gradient descent method}
In this study, the gradient descent method is employed for path finding. Gradient descent is an
iterative mathematical optimization algorithm utilized to locate the minimum of a function. It
involves taking steps proportional to the negative gradient (or estimated gradient) of the
function at the current point. If steps in the positive gradient direction are taken, the algorithm
approaches the maximum of the function, known as the incremental gradient process.
Here, the function of interest is discrete and two-dimensional, derived as a map from the A*
method. The following pseudocode outlines the process of estimating the gradient on this
map and determining the path:

Due to the heuristic criterion employed in the $A^*$ algorithm, only the map regions, where the
presence of a path is more probable, are valued. Consequently, a potential error arises at the
border between the valued area and the non-valued area when using the decreasing gradient
method. This issue is addressed by adding the fourth and fifth lines, thereby rectifying the
error.

\subsection{Proposed algorithm for local planning}

After establishing the global path from the origin to the destination, accounting for static
obstacles, the next step involves local planning to fine-tune the final trajectory for the vehicle.
In this research, the local trajectory is defined as a sequence of vehicle positions along with
the corresponding time intervals. Each position is characterized by four parameters detailing
the vehicle's location and orientation.

\begin{algorithm}[H]
\caption{Pseudocode for the modified $A^*$ Algorithm}
\label{Alg1}
\begin{algorithmic}[1]
\State Initialize the overall cost-map with the start and goal locations.
\State Form an array of ordered pairs, where the first component denotes the cell index and the second component represents the cell value.
\State Set the initial value of all map points (except the starting point, which is set to zero) to a large value.
\State Place the starting point with a value of zero in the first position of the queue.
\While{the queue is not empty}
    \State Extract the cell with the highest value from the queue.
    \If{the end point is reached}
        \State Terminate the process.
    \EndIf
    \ForAll{neighboring cells (left, right, top, bottom)}
        \If{cell value is less than a predefined threshold}
            \State Remove the cell from consideration.
        \EndIf
        \If{cell is outside the map boundary}
            \State Remove the cell from consideration.
        \EndIf
        \State Calculate the new value for the cell.
        \State Compute the Euclidean distance from the cell to the end point.
        \State Store the cell index, its new value, and the sum of the new value.
    \EndFor
\EndWhile
\end{algorithmic}
\end{algorithm}

\begin{algorithm}[H]
\caption{Pseudocode for $A^*$ Algorithm}
\label{Alg2}
\begin{algorithmic}[1]
\State Begin from the destination point.
\State Continue until reaching the starting point:
\While {not at starting point}
    \If {each neighboring cell value = $POT\_HIGH^*$}
         \State Move to the neighboring cell with the lowest value.
    \Else
         \State Compute the numerical approximation of the potential gradient.
        \If {gradient is zero}
             \State An error has occurred.
        \EndIf
         \State Move half a cell in the direction of the negative gradient.
    \EndIf
\EndWhile 
\end{algorithmic}
\footnotesize $^*POT\_HIGH$: A very high initial value
\end{algorithm}

\subsection{Setting the trajectory density}

The proposal outlined in this article aims to enhance trajectory quality while mitigating
computational burden by augmenting the number of waypoints in critical sections, such as
bends or turns. This is achieved by integrating a dynamic term into the time intervals within
the fastest time cost function, enabling trajectory accuracy adjustments along specific
segments. The cost function for the fastest route is introduced as Eq. \ref{eq:formula1}.

\begin{minipage}{0.45\linewidth}
\begin{equation}
f_k = \sum_{k=1}^{n-1} \left( \Delta T_k \right)^2
\label{eq:formula1}
\end{equation}
\end{minipage}
\hfill
\begin{minipage}{0.45\linewidth}
\begin{equation}
f_k = \sum_{k=1}^{n-1} w_k \left( \Delta T_k \right)^2
\label{eq:formula2}
\end{equation}
\end{minipage}

By selecting this cost function and employing the Lagrange coefficient, the inclination is to
establish uniform time intervals throughout the path. However, assigning specific coefficients
to individual time intervals allows for customization of their durations. Thus, the revised cost
function is as Eq.\ref{eq:formula2}.





\begin{figure}
  \centering
  \includegraphics[width=0.2\textwidth]{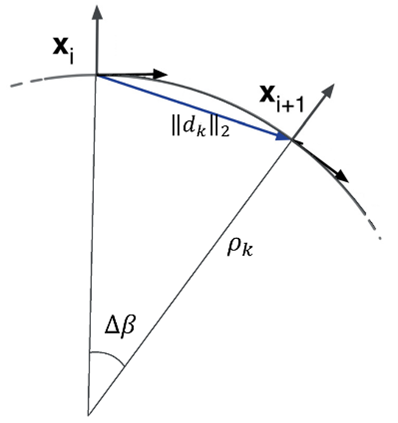}
  \caption{Calculation of path curvature}
  \label{fig:PathCurvature}
\end{figure}

The larger the $w_k$ weight is, the smaller its corresponding time interval $\Delta T_k$ will be, and as a
result, the accuracy of the trajectory will increase. Since the turns and curves of the route are
among its sensitive parts\cite{li2022autonomous}, in this research, the accuracy of the path has been increased in the
turns and curves of the route, and in the parts where the car moves in a straight line, in order
to reduce the computational burden, a smaller number of waypoints have been used. The
Eq.\ref{eq:formula3} is used to identify the path curves, which is obtained according to Fig.\ref{fig:PathCurvature}.

\begin{equation}
\rho_k = \frac{\|d_k\|_2}{|2\sin\left(\frac{\Delta\beta_k}{2}\right)|} \quad (\text{since } \Delta\beta_k \ll 1) \quad \rho_k = \frac{\|d_k\|_2}{|\Delta\beta_k|} \geq \rho_{\text{min}}
\label{eq:formula3}
\end{equation}

This equation gives the radius of curvature of the path, the smaller the radius of curvature of
the path, the more winding the path is. Therefore, according to the equation (4), the inverse
value obtained from (3) is considered as the weight and is used for $w_k$ in the (5).

\begin{equation}
w_k = \frac{|\Delta\beta_k|}{\|d_k\|_2} 
\label{eq:formula4}
\end{equation}

\subsection{Obstacle dynamics in local planning}
For dynamic obstacle tracking, the position of the obstacle center is calculated every time the
cost map is updated and given to the Kalman filter. A critical difference between motion
planning for vehicles and mobile robots lies in the nature of the environment they navigate. In
the case of vehicles, the planning environment encompasses fast moving vehicles,
necessitating consideration of accelerated obstacle movement models. The Kalman filter
emerges as a robust tool for providing scientific and engineering predictions regarding the
future states of dynamic systems, particularly in scenarios where information about the
system is imprecise. Notably, the Kalman filter boasts efficiency, requiring minimal memory
as it relies solely on past state information. In this study, the Kalman filter leverages a
constant acceleration motion model to estimate obstacle movement, thus yielding the
following dynamic system equation.

\begin{equation}
\text{Pos\_est} = \text{Pos\_old} + V_{\text{rel}} \Delta t + \frac{1}{2} A_{\text{rel}} \Delta t^2 
\label{eq:formula5}
\end{equation}

where, \text{Pos\_est} is the estimated position of the obstacle, \text{Pos\_old} is the previous position of the
obstacle, $V_{\text{rel}}$ is the relative speed of the obstacle, $A_{\text{rel}}$ is the obstacle acceleration and $\Delta t$
represents the time interval between both iterations of the algorithm. To estimate the position,
speed and acceleration of the obstacle, the relationship between these three parameters should
be written in a standard way. The following equations contain these relationships.

\begin{align}
\begin{bmatrix}
\text{Pos}_{n+1} \\
\text{Vel}_{n+1}
\end{bmatrix}
&= 
\begin{bmatrix}
1 & t \\
0 & 1
\end{bmatrix}
\begin{bmatrix}
\text{Pos}_n \\
\text{Vel}_n
\end{bmatrix}
+
\begin{bmatrix}
\frac{t^2}{2} \\
t
\end{bmatrix}
\text{Acc}_n + \mathbf{w}_n  \\
Z_{n+1}
&= 
\begin{bmatrix}
1 & 0
\end{bmatrix}
\begin{bmatrix}
\text{Pos}_{n+1} \\
\text{Vel}_{n+1}
\end{bmatrix}
+ v_{n+1}
\end{align}

$w_n$ is process noise and $v_n+1$ is observation noise, which are considered as white with zero mean. Considering that the above equation has an uncertain input of acceleration, the following equations can be used.

\begin{align}
\begin{bmatrix}
\text{Pos}_{n+1} \\
\text{Vel}_{n+1} \\
\text{Acc}_{n+1}
\end{bmatrix}
&= 
\begin{bmatrix}
1 & t & \frac{t^2}{2} \\
0 & 1 & t \\
0 & 0 & 1
\end{bmatrix}
\begin{bmatrix}
\text{Pos}_n \\
\text{Vel}_n \\
\text{Acc}_n
\end{bmatrix}
+ \mathbf{w}_n \\
Z_{n+1}
&= 
\begin{bmatrix}
1 & t & \frac{t^2}{2}
\end{bmatrix}
\begin{bmatrix}
\text{Pos}_n \\
\text{Vel}_n \\
\text{Acc}_n
\end{bmatrix}
+ \mathbf{w}_n + v_{n+1}
\end{align}

Table 1 shows the necessary definitions to use in Kalman equations.

\begin{table*}[htbp]
\centering
\caption{System Augmented State}
\label{tab:system_aug_state}
\begin{tabular}{|l|l|}
\hline
System state vector                  & $x_{\text{aug}}(n) = \begin{bmatrix} \text{Pos}_n \\ \text{Vel}_n \\ \text{Acc}_n \end{bmatrix}$ \\ \hline
State transition model matrix       & $F_{\text{aug}}(n) = \begin{bmatrix} 1 & t & \frac{t^2}{2} \\ 0 & 1 & t \\ 0 & 0 & 1 \end{bmatrix}$ \\ \hline
Effect of noise level        & $G_{\text{aug}}(n) = \begin{bmatrix} 1 \\ 1 \\ 1 \end{bmatrix}$ \\ \hline
Observation model vector            & $H_{\text{aug}}(n) = \begin{bmatrix} 1 & t & \frac{t^2}{2} \end{bmatrix}$ \\ \hline
Observation noise                   & $v_{\text{aug}}(n) = w_n + v_{n+1}$ \\ \hline
Process noise covariance            & $Q_{\text{aug}}(n) = E\{w_{\text{aug}}(n) w_{\text{aug}}^T(n)\} = Q(n)$ \\ \hline
Covariance of observation noise     & $R_{\text{aug}}(n) = E\{v_{\text{aug}}(n) v_{\text{aug}}^T(n)\}$ \\
                                    & \quad\quad $= H(n)G(n)Q(n)G^T(n)H^T(n) + R(n)$ \\ \hline
Correlation matrix of process noise & $T_{\text{aug}}(n) = E\{w_{\text{aug}}(n) v_{\text{aug}}^T(n)\} = Q(n)G^T(n)H^T(n)$ \\ \hline
\end{tabular}
\end{table*}

The following equations show the Kalman relations necessary to calculate the speed.

\begin{multline}
\hat{x}{\ }_{\text{aug}}(n|n-1) = F_{\text{aug}}(n) \hat{x}{\ }_{\text{aug}}(n-1|n-1) \\
\mathrm{\Sigma}_{\text{aug}}(n|n-1) = F_{\text{aug}}(n) \mathrm{\Sigma}_{\text{aug}}(n-1|n-1) F_{\text{aug}}^T(n) \\
+ G_{\text{aug}}(n) Q_{\text{aug}}(n) G_{\text{aug}}^T(n) \\
y\hat{(n)} = Z_{\text{aug}}(n) - H_{\text{aug}}(n) \hat{x}{\ }_{\text{aug}}(n|n-1) \\
K_{\text{aug}}(n) = [\mathrm{\Sigma}_{\text{aug}}(n|n-1) H_{\text{aug}}^T(n) \\
+ G_{\text{aug}}(n) T_{\text{aug}}(n)] R_{\text{aug}}^{-1}(n) \\
\hat{x}{\ }_{\text{aug}}(n|n) = \hat{x}{\ }_{\text{aug}}(n|n-1) + K_{\text{aug}}(n)y\hat{(n)} \\
\mathrm{\Sigma}_{\text{aug}}(n|n) = \mathrm{\Sigma}_{\text{aug}}(n|n-1) \\
- K_{\text{aug}}(n) H_{\text{aug}}(n) \mathrm{\Sigma}_{\text{aug}}(n|n-1)
\end{multline}

In the above equations, $\hat{x}_{\text{aug}}$ is the estimated state vector, $\Sigma_{\text{aug}}$ is the covariance matrix of the estimation error, and $K_{\text{aug}}$ is the Kalman gain. 

How to design the covariance noise process (Q matrix) is explained below. In practice, a lot
of time is spent simulating and evaluating the collected data to choose the right value for Q.
In general, the process model will be in the following form:

\begin{minipage}{0.45\linewidth}
\begin{equation}
\dot{x} = Ax + Bu + w
\end{equation}
\end{minipage}
\hfill
\begin{minipage}{0.45\linewidth}
\begin{equation}
f(x) = Fx + \Gamma w 
\end{equation}
\end{minipage}

where $w$ is the process noise.

The desired dynamic system is modeled using position, speed, and acceleration. Now it is assumed that the acceleration, which is larger than the order, is constant in specific time intervals that are independent of each other and changes at the end of each interval. In other words, the acceleration jumps in each time interval and is modeled as below:

\begin{minipage}{0.45\linewidth}
\begin{equation}
F = \begin{bmatrix}
1 & t & \frac{t^2}{2} \\
0 & 1 & t \\
0 & 0 & 1
\end{bmatrix}
\end{equation}
\end{minipage}
\hfill
\begin{minipage}{0.45\linewidth}
\begin{equation}
\Gamma = \begin{bmatrix}
\frac{\Delta t^2}{2} \\
\Delta t \\
1
\end{bmatrix}
\end{equation}
\end{minipage}

where $\Gamma$ is the system noise gain and $w$ is the desired continuous piece acceleration. The transfer matrix of the system is also defined as follows.

Therefore, the covariance matrix of the system will be as follows.

\begin{minipage}{0.45\linewidth}
\begin{equation}
Q = \mathbb{E}[\Gamma w(t) w(t) \Gamma^T] = \Gamma \sigma_v^2 \Gamma^T
\end{equation}
\end{minipage}
\hfill
\begin{minipage}{0.45\linewidth}
\begin{equation}
Q = \begin{bmatrix}
\frac{\Delta t^4}{4} & \frac{\Delta t^3}{2} & \frac{\Delta t^2}{2} \\
\frac{\Delta t^3}{2} & \Delta t^2 & \Delta t \\
\frac{\Delta t^2}{2} & \Delta t & 1
\end{bmatrix}
\end{equation}
\end{minipage}

\section{Experiments}

\begin{figure}[]
  \centering
  \includegraphics[width=\linewidth]{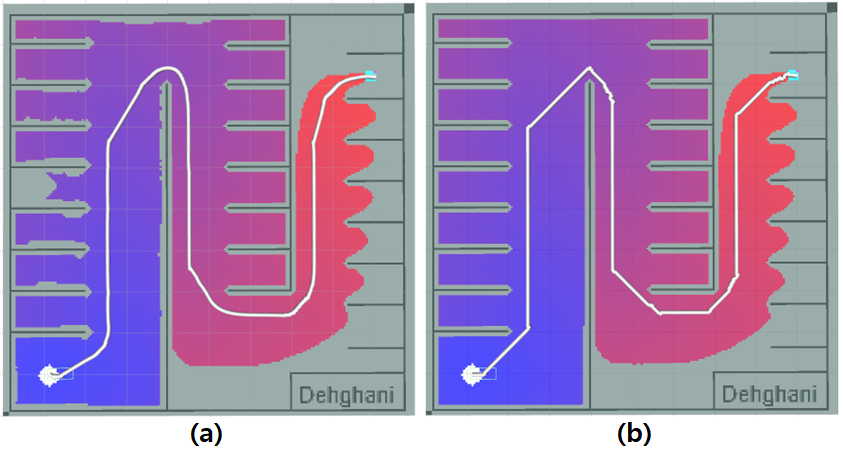}
  \caption{ Comparison of path planning with a) descent gradient and b) normal $A^*$}
  \label{fig:smootherpath}
\end{figure}

Fig.\ref{fig:smootherpath} illustrates the scenario involving a parking lot, showcasing two planning methods: normal $A^*$ and gradient descent. Comparing the paths generated by these methods reveals that the gradient descent approach yields a notably smoother and optimal path compared to the conventional $A^*$ method. It's worth noting that while the global planning may exhibit suboptimal outcomes, the local planner effectively mitigates many of its drawbacks, this means that the imperfections are acceptable and do not need special care when putting the plan into action.
In designing the global planner, the paramount considerations include ensuring completeness, accuracy, and reducing computational complexity. Simulations validate that the proposed algorithm in this study satisfactorily fulfills these criteria, making it well-suited for global path planning.

\begin{figure}[]
  \centering
  \includegraphics[width=\linewidth]{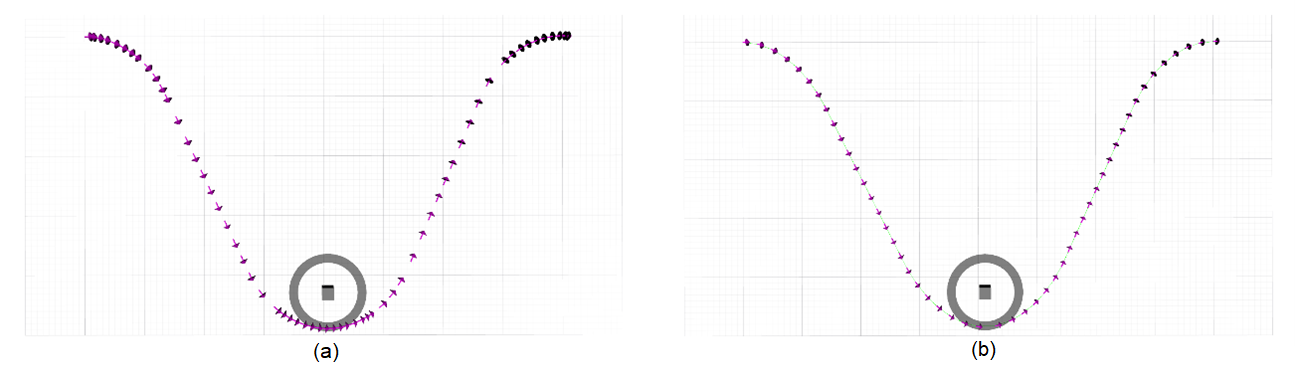}
  \caption{a) normal path planning b) considering density in the bends along the path}
  \label{fig:trajectory density}
\end{figure}

Fig.\ref{fig:trajectory density} illustrates the impact of path density enhancement. Here, the origin is located at (-4,
0) and the destination at (+4, 0), while the obstacle has shifted from a position above the
horizontal axis to (0, -4.5). Each arrow along the path indicates specific positions and
directions for the car to traverse towards reaching the destination. Despite local optimization
efforts and the path's continuous shape alteration, without multipath optimization, the
resulting path remains suboptimal. This scenario was conducted solely to introduce curvature
into the path and evaluate improvements in this aspect. Higher curvature regions entail more
waypoints, thus increasing path density accordingly.

To assess the effectiveness of the Kalman filter, various scenarios were examined involving
moving obstacles with diverse velocities and accelerations. Their positions were calculated,
and this data was subsequently fed into the Kalman filter to extract the motion model of the
obstacle.

\subsection{First scenario: The obstacle moves at a constant speed in the vertical (y) direction}

\begin{figure*}
    \centering
    \subfigure[]{\includegraphics[width=0.3\textwidth]{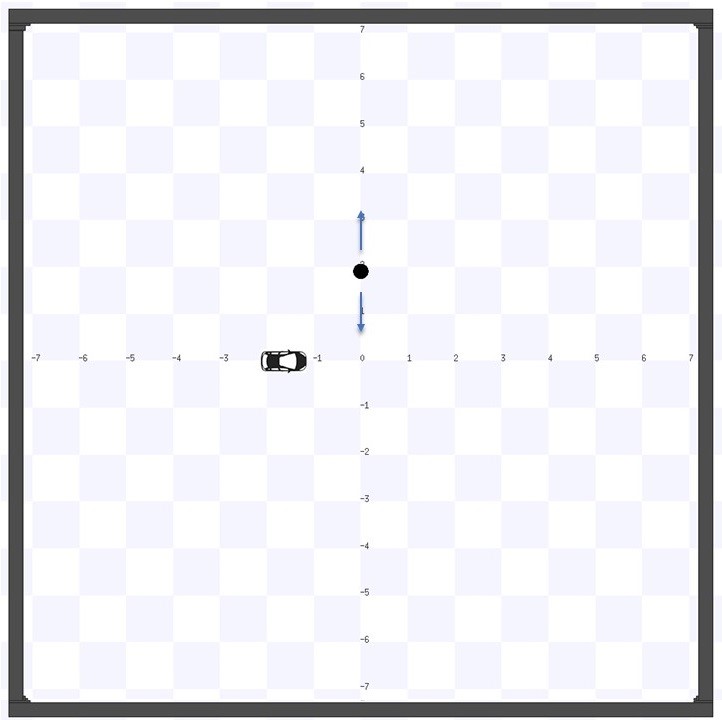}} 
    \subfigure[]{\includegraphics[width=0.3\textwidth]{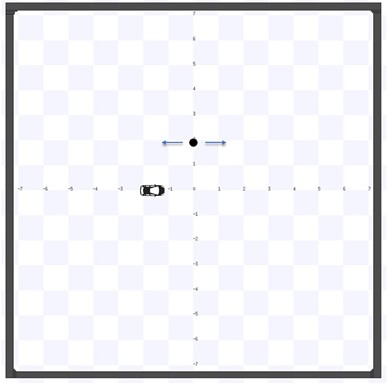}} 
    \subfigure[]{\includegraphics[width=0.3\textwidth]{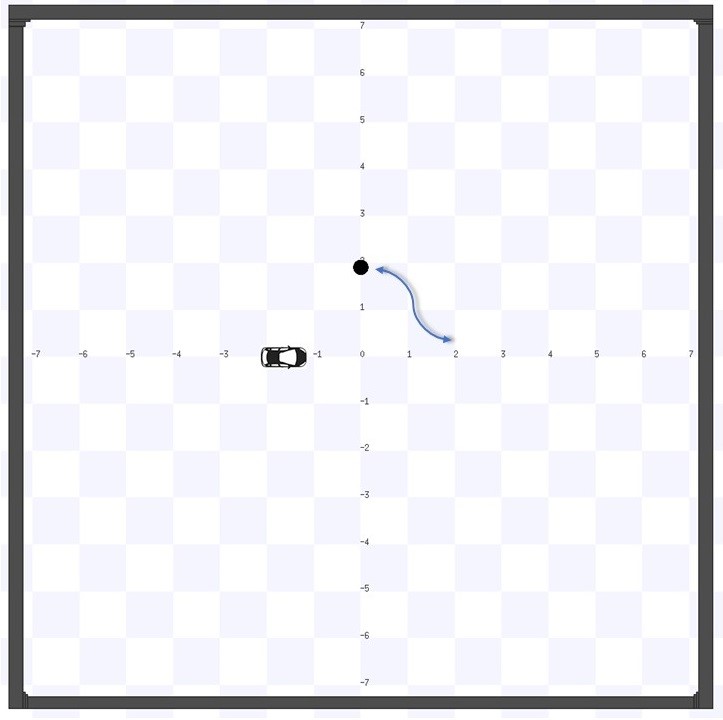}}\\
    \subfigure[]{\includegraphics[width=0.3\textwidth]{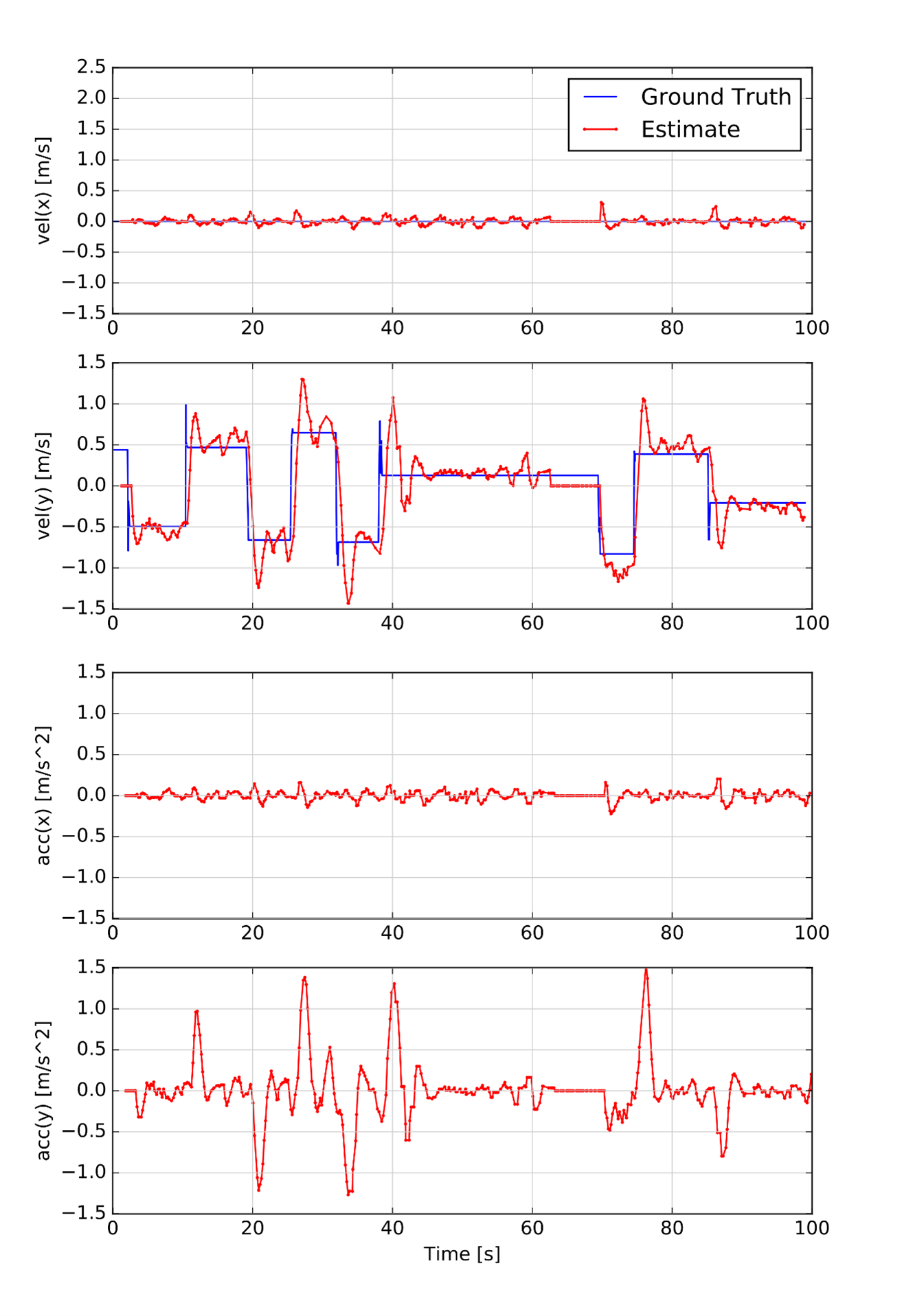}} 
    \subfigure[]{\includegraphics[width=0.3\textwidth]{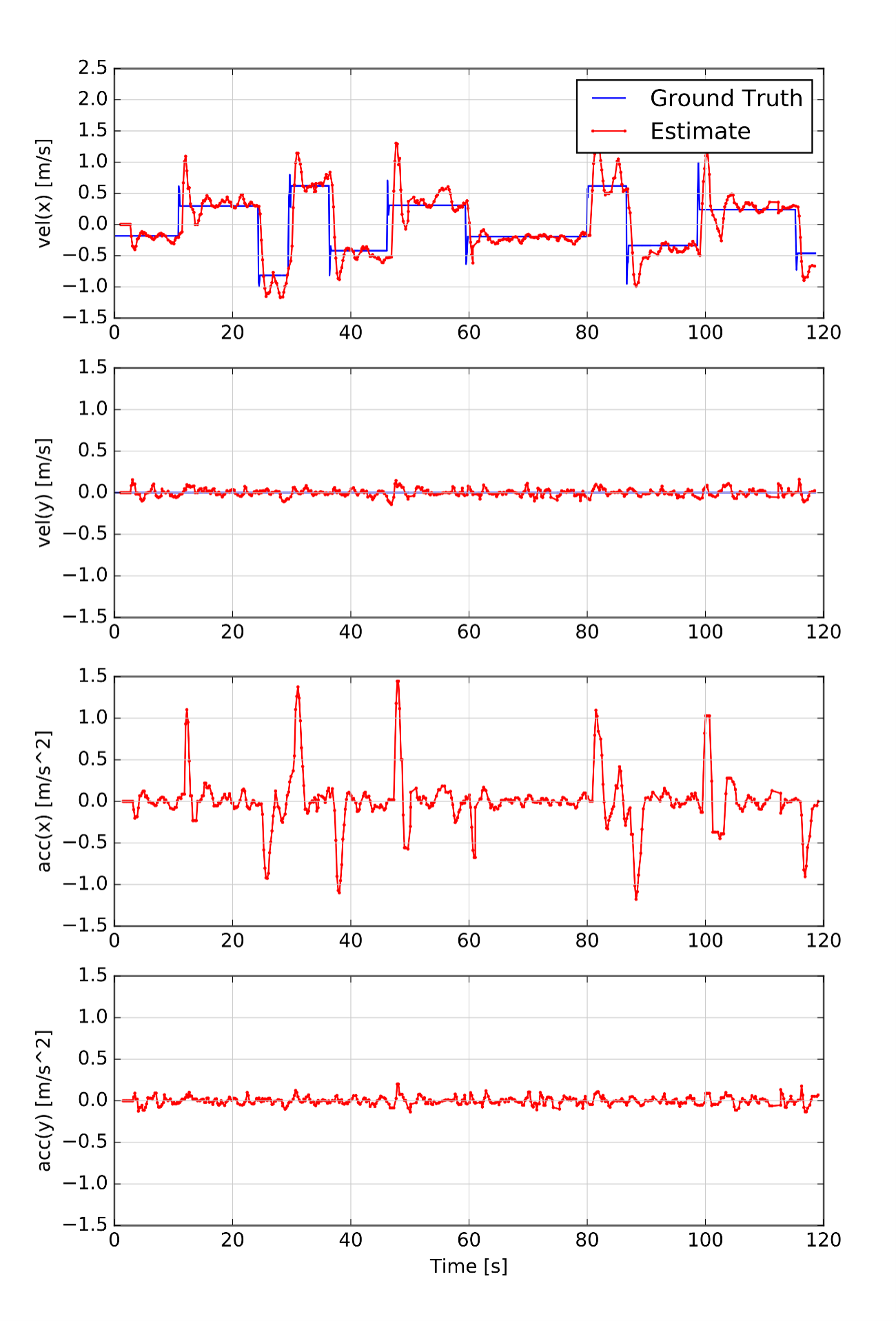}} 
    \subfigure[]{\includegraphics[width=0.3\textwidth]{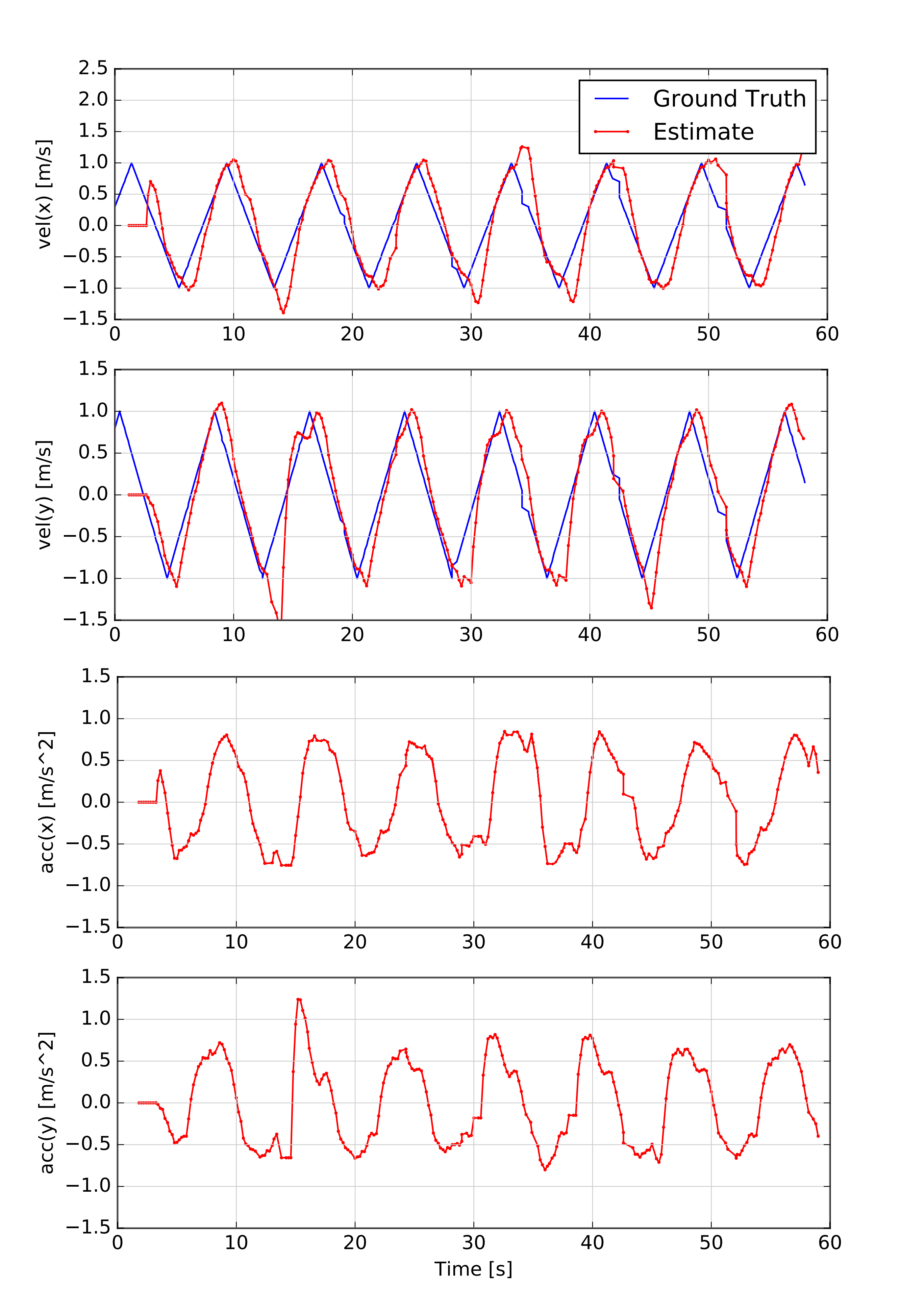}}
    \caption{(a) Path planning in the presence of an obstacle moving in the y direction (b) Evaluation of the Kalman filter for the moving obstacle in the y
direction (c) moving obstacle in the x direction (d) Kalman filter for the moving obstacle with constan (e) moving obstacle with different constant accelerations (f) Evaluation of the Kalman filter for an accelerated moving obstacle }
    \label{fig:scenarios}
\end{figure*}

In this scenario, the moving obstacle is positioned ahead of the vehicle and travels along the
y-axis (Fig.\ref{fig:scenarios}(a)). The obstacle's displacement is such that its linear velocity in the x-axis is zero, while
its linear velocity in the y-axis changes completely randomly.

To confine the obstacle within the planning environment, its movement is constrained to a
defined interval, with its direction changed upon reaching upper and lower limits. Despite
large error in the perception phase and challenges in calculating the obstacle's position, the
Kalman filter adeptly estimates the speed and acceleration of the obstacle system.(Fig.\ref{fig:scenarios}(d))

\subsection{Second scenario: The obstacle moves at a constant speed in the vertical (x) direction}

In this scenario, the obstacle is positioned ahead of the vehicle and travels along the x-axis.(Fig.\ref{fig:scenarios}(b))
Its displacement is designed so that its linear velocity in the y-direction remains zero while its
linear velocity in the x-direction varies randomly. To ensure the obstacle remains within the
routing environment, its movement is confined to a specific interval, and its direction is
altered upon reaching the left and right boundaries. Despite significant error in calculating the
obstacle's position, the Kalman filter effectively estimates the speed and acceleration of the
obstacle system.(Fig.\ref{fig:scenarios}(e)) This scenario was tested to demonstrate the algorithm's robustness to
variations in the obstacle's movement direction.

\subsection{Third scenario: The obstacle moves with a constant acceleration}

To assess the Kalman filter's performance in scenarios involving accelerated motion, a
moving obstacle was subjected to constant acceleration within the environment. To limit the
obstacle's movement within the designated area, its linear velocities in the $x$ and $y$ directions
were modulated using triangular functions. Depending on the frequency and amplitude of
these functions, the trajectory of the obstacle varied. Despite the challenges in accurately
calculating the obstacle's position, the Kalman filter effectively estimated its speed and
acceleration, highlighting its robustness even in the presence of significant error in the
perception phase.(Fig.\ref{fig:scenarios}(c)(f))

\section{Motion planning in the presence of moving obstacles}
    
    \subsection{First test: Static obstacles}
    
    In order to test the performance of the algorithm, the initial values specified in Table \ref{tab:combined_scenario} have
    been used.

    
    
\begin{figure*}
    \centering
    \subfigure[]{\includegraphics[width=0.3\textwidth]{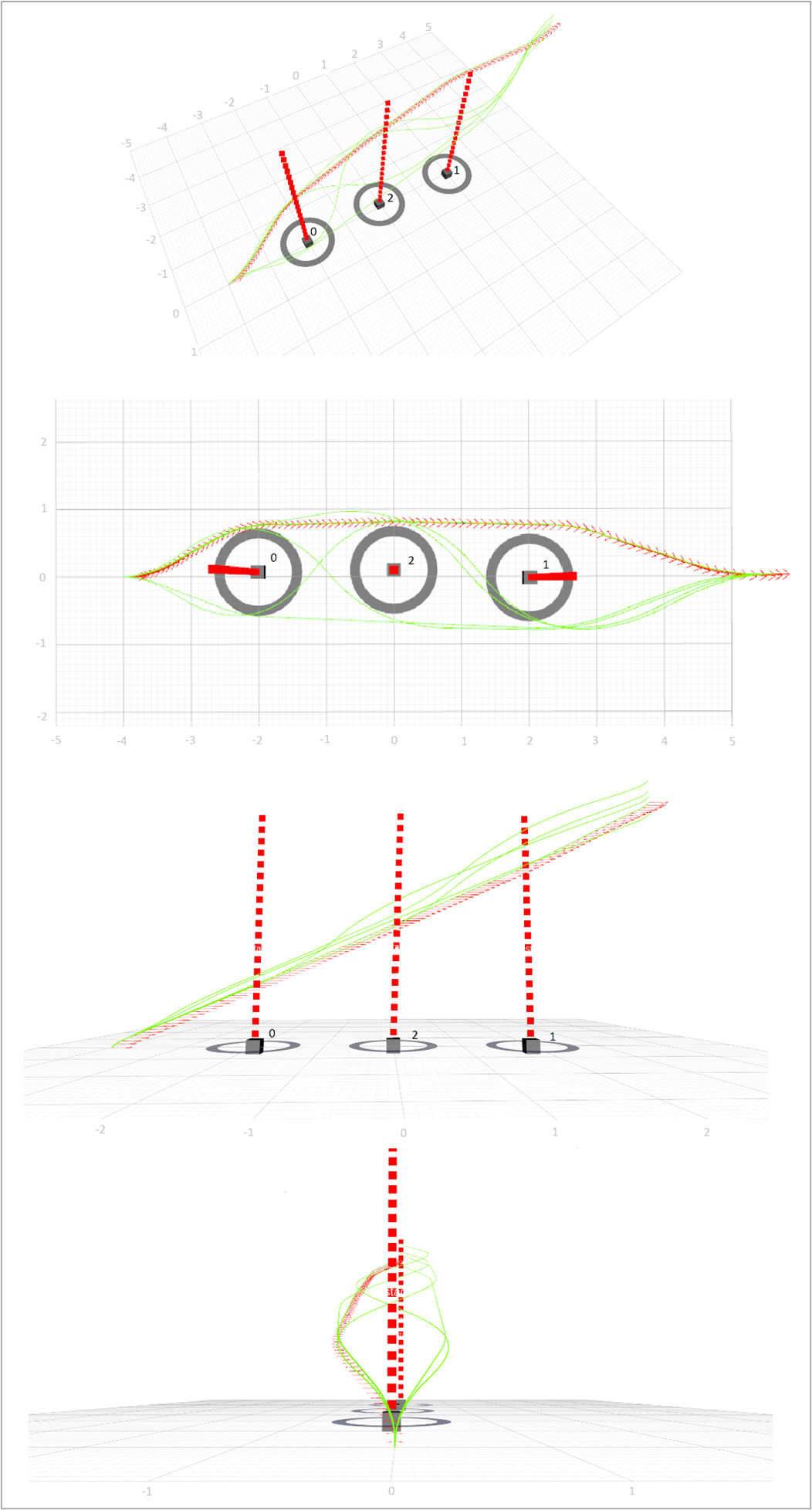}} 
    \subfigure[]{\includegraphics[width=0.3\textwidth]{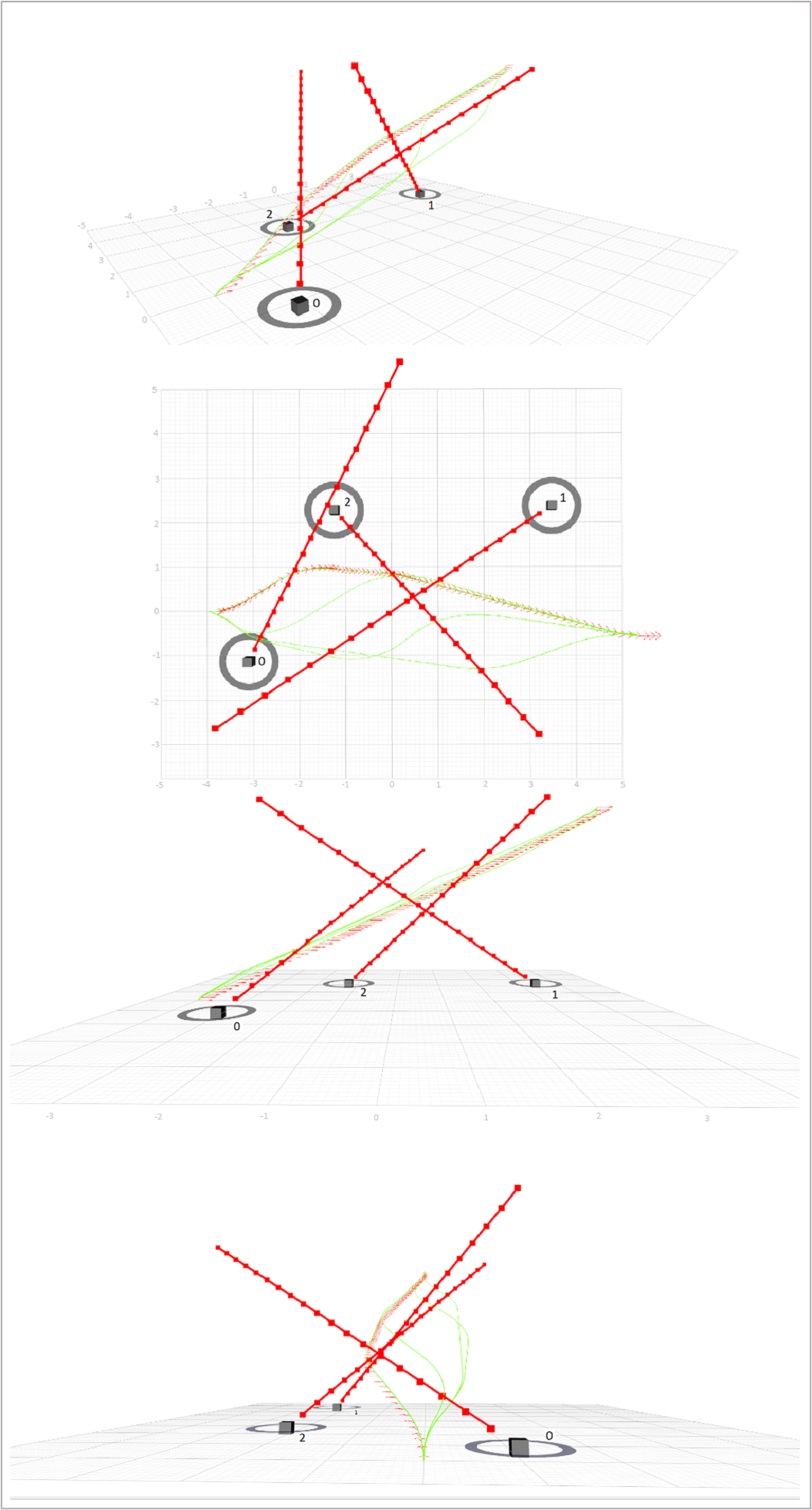}} 
    \subfigure[]{\includegraphics[width=0.3\textwidth]{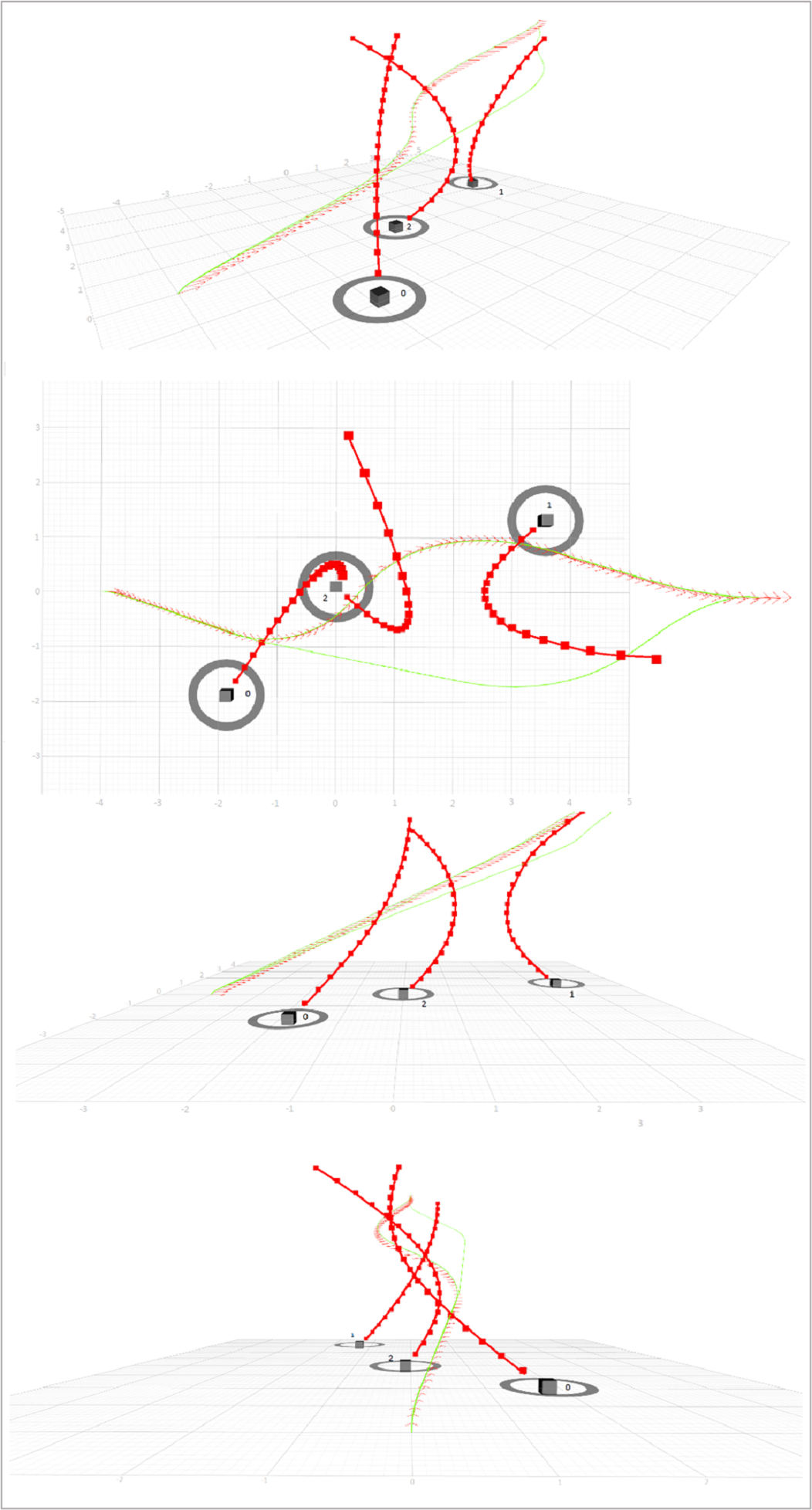}}
    \caption{(a) Two-dimensional position over time, perspective view, from above,
from the front, and from the left in the presence of static obstacles (b) in the presence of constant speed obstacles (c) in the presence of constant acceleration
obstacles}
    \label{fig:fig5}
\end{figure*}

    The maximum number of homotopy classes is limited at 5 to control the computational
    workload. Consequently, 5 trajectories are simultaneously generated, and the one with the
    lowest cost is chosen. In the fig.\ref{fig:fig5}(a) red arrows denote the waypoints for the vehicle to traverse from
    the origin to the destination. The obstacles are considered as points and the circle around
    them indicates the safety margin. Obstacles can have any shape and be placed in any position.
    In order to show how the obstacles affect the path, it has been tried to place the obstacles at a
    point that has the greatest impact on the path. The final trajectory comprises 81 states, with an
    estimated time of arrival at 20.14 seconds. The average path computation time is 14.91milliseconds, which is considered efficient for a computer equipped with a dual-core
    processor running at 1.8 GHz. The green paths along by the optimal trajectory are homotopy paths.

\begin{table}[htbp]
\centering
\caption{Initial conditions for all three tests}
\label{tab:combined_scenario}
\resizebox{\textwidth}{!}{%
\begin{tabular}{|>{\centering\arraybackslash}p{4cm}|>{\centering\arraybackslash}p{3cm}|>{\centering\arraybackslash}p{4cm}|>{\centering\arraybackslash}p{4cm}|}
\hline
\textbf{Description}                & \textbf{Static Obstacle} & \textbf{Obstacle with Constant Speed} & \textbf{Obstacle with Constant Acceleration} \\ \hline
Initial position of the vehicle     & \multicolumn{3}{c|}{$[-4, 0]$} \\ \hline
Destination location of the vehicle & \multicolumn{3}{c|}{$[4, 0]$} \\ \hline
Initial position of obstacles       & $0:[-2, 0]$ & $1:[2, 0]$ & $2:[0, 0]$ \\ \hline
Initial speed of obstacles          & $0:[0, 0]$  & $1:[-0.2, -0.3]$ & $2:[0.2, -0.2]$ \\ \hline
Initial acceleration of obstacles   & $0:[0, 0]$  & $1:[0, 0]$ & $2:[-0.02, 0.03]$ \\ \hline
Dynamic conditions of obstacles     & Completely static & Constant velocity & Constant acceleration \\ \hline
\end{tabular}%
}
\end{table}

\subsection{Second scenario: moving obstacles at a constant speed}

In order to test the performance of the algorithm, the initial values specified in Table \ref{tab:combined_scenario} have been used in this scenario.



The maximum number of homotopy classes is set to four to avoid the increase in the
computational cost. Therefore, four trajectories are designed simultaneously and the one that
has a lower cost is selected. In the fig.\ref{fig:fig5}(b) the red arrows show the situations that the car must go through
to go from the origin to the destination. The obstacles are simplified to points, with a
surrounding circle denoting the safety margin. These obstacles are versatile, capable of
assuming any shape and position. To demonstrate their influence on the path, obstacles are
strategically positioned to show their maximum impact.
The resulting route includes 75 states and the time to reach the destination will be 19.22
seconds. The average time required to calculate the path is 23.1 milliseconds, which is a good
time for a computer with a dual-core processor with a frequency of 1.8 GHz.

\subsection{Third test: Accelerated moving obstacles}
In order to test the performance of the algorithm, the initial values specified in Table \ref{tab:combined_scenario} have been used.

 



The resulting route includes 85 states and the time to reach the destination will be 21.12
seconds. The average time required to calculate the route is 27.32 seconds, which is a good
time for a computer with a dual-core processor with a frequency of 1.8 GHz. Fig.\ref{fig:fig5}(c) illustrates the scenario of accelerated obstacles.

\section{Conclusion}

This research introduces the novel concept of "Trajectory density" to assess the quality of generated paths by vehicle motion planning algorithms. By defining a new objective function and applying dynamic coefficients, this criterion is enhanced. Given that increasing the number of track conditions escalates computational load, it's impractical to boost trajectory density throughout its entirety. Hence, the technique proposed in this study detects sensitive areas of the route, such as bends, and adjusts the density accordingly. Motion planning in the proposed method incorporates dynamics of moving obstacles. To identify obstacles and their locations, a new method is employed, amalgamating sensor data to produce a local cost-map.
Computer vision methods are then utilized to differentiate between fixed and moving
obstacles, with the latter's location provided at a specific frequency enabling dynamic obstacle tracking. This information is integrated into the Kalman filter to estimate speed and acceleration, enabling extraction of obstacle dynamics.



{\small
\bibliographystyle{splncs04}
\bibliography{Paper}
}





\end{document}